\documentclass[11pt,a4paper]{article}
\usepackage[margin=0.74in]{geometry}
\usepackage[T1]{fontenc}
\usepackage[utf8]{inputenc}
\usepackage{lmodern,microtype}
\usepackage{amsmath,amssymb,booktabs,array,graphicx,caption,subcaption}
\usepackage[colorlinks=true,urlcolor=blue,citecolor=blue,linkcolor=blue]{hyperref}
\usepackage{natbib}
\hypersetup{pdftitle={When Knowledge-Tracing Rankings Depend on Available History},pdfauthor={Devis Wawan Saputra}}
\title{When Knowledge-Tracing Rankings Depend on Available History\\\large A Source-Correction and Matched-Information Audit on ASSISTments 2009--2015}
\author{Devis Wawan Saputra\\\small Independent Researcher, Indonesia\\\small ORCID: \href{https://orcid.org/0000-0002-7133-9410}{0000-0002-7133-9410}}
\date{7 November 2026}
\begin{document}
\maketitle
\begin{abstract}
Knowledge-tracing benchmarks often compare models that have different access to a learner's previous responses. We examine how data provenance, history length, and representation affect those comparisons. First, we compare a pinned ASSISTments 2009 sequence mirror against the corrected official release. Of 4,148 learners represented in both, 817 have different ordered skill--response sequences; the mirror also contains fewer labeled events. Second, we retain earlier learner-disjoint model evaluations on the 2009 and 2015 datasets, distinguishing exploratory results from a corrective 2015 replication. Finally, we evaluate a new, nonoverlapping sample of 4,000 ASSISTments 2015 learners using three models constrained to the same one-event or 20-event prior-response window. On 20,592 test responses from 582 learners, the 20-event Brier scores were 0.17667 for count-based logistic regression, 0.17389 for an order-aware logistic model, and 0.17514 for a GRU. The GRU-minus-order-aware difference was 0.00125 (95\% learner-cluster bootstrap interval [0.00008, 0.00243]); at one prior event, the corresponding difference was 0.00212 [0.00120, 0.00301]. Earlier comparisons had favored the GRU against simpler history summaries. The newer analysis instead favors a logistic comparator that represents order using the same permitted event stream. These results concern predictive performance on particular historical cohorts and model settings. They neither isolate an architectural effect nor establish generalization to contemporary classrooms.

\end{abstract}

\section{Introduction}
Knowledge tracing predicts a student's next response from earlier practice. Binary correctness histories allow comparisons among simple frequency estimates, probabilistic mastery models, logistic models, and neural sequence encoders. Such comparisons are useful only if the underlying event records and the information supplied to each predictor are described accurately.

Two problems recur in published benchmarks. Datasets distributed under the same name may reflect different filtering and serialization decisions. Moreover, a recurrent predictor may see answers on other skills and their order, while a conventional performance-factors model sees counts for the queried skill. A performance difference in that setting combines model specification with differences in available history. It cannot be attributed to recurrence alone.

We investigate these problems using the ASSISTments 2009--2010 and 2015 Skill Builder releases. We first document what changes between a widely used transformed 2009 mirror and the source-corrected official records. We then retain the earlier held-out results, including a failed optimization diagnostic and its independently sampled follow-up. The main new experiment restricts all models to the same prior-response window and scores their predictions on precisely the same held-out response events. We ask whether the GRU advantage reported against a count-based baseline persists after adding explicit order information to an interpretable logistic model, and whether that comparison changes between windows of one and 20 previous events.

\section{Related Work and Positioning}
Bayesian Knowledge Tracing (BKT) represents learning through an evolving latent mastery probability \citep{corbett1994}. Performance Factors Analysis (PFA) uses previous successes and failures as predictors in a logistic model \citep{pavlik2009}; Deep Knowledge Tracing introduced recurrent sequence encoders \citep{piech2015}. Later studies and surveys found that comparisons vary with datasets, splits, preprocessing, and baseline implementation \citep{gervet2020,abdelrahman2023}.

This is an active methodological literature. A 2026 replication package compares BKT, PFA, and neural methods using ASSISTments cohorts \citep{polydoros2026}. Padi and colleagues test information-matched prediction and identify leakage associated with duplicated skill records \citep{padi2026}. Badran and Preisach examine leakage-free and recency-aware sequence representations \citep{badran2026}. Thus, correcting historical data and equalizing input information are not novel ideas by themselves. Our narrower test is a frozen history-window comparison on a new 2015 learner sample, reported alongside a concrete check of the 2009 mirror's provenance. We do not claim a new knowledge-tracing architecture or a comprehensive ranking of existing models.

\section{Study Design and Evidence}
The results come from three related but distinct analyses. The original 2009 study reproduces six models on a fixed learner-level split of a transformed mirror. Additional 2009 analyses vary PFA features and GRU training; since their test set had already been inspected, these are exploratory. A separate audit compares the mirror with the corrected official source and refits PFA and GRU to that source.

The 2015 study proceeded in stages. A first 5,000-learner sample produced a nonconverged PFA fit. A second, nonoverlapping 5,000-learner sample used a corrected solver and provided a new held-out comparison. After those results were available, we fixed the protocol for a third sample of 4,000 learners, excluding everyone from the first two samples. That third analysis, comparing one- and 20-event history windows, is the present paper's central methodological experiment. Its design was recorded before examining the third sample's results, although the choice of research question followed earlier analyses and is not a public preregistration.

\section{Data and Evaluation Protocol}
\subsection{Dataset origin and observed differences from the corrected source}
The intended underlying source is the ASSISTments 2009--2010 Skill Builder data documented by its maintainers \citep{assistmentsdata}; the ASSISTments tutoring platform and its formative-assessment context were described by \citet{feng2009}. The execution pipeline reads \texttt{Atomi/ASSISTments2009} at pinned revision \texttt{c72a664a9693547fb206652}\allowbreak\texttt{ed2ce18e62d320c7d} and file \path{data/train-00000-of-00001.parquet}. The archived file SHA-256 is \texttt{2053b7ea63f1d5ea953834bf82ed8954}\\\texttt{3e1f11bae1e3e7535c2f9d7f49d4dda7}. The mirror contains one serialized sequence per learner. The empirical record reports 4,148 unique learners and 274,331 binary response observations.

\textbf{Direct provenance audit.} The ASSISTments maintainers corrected an earlier Skill Builder release after identifying duplicated records \citep{assistmentsdata}. We downloaded the official corrected student--problem CSV directly from the file they link (SHA-256 \texttt{162ef8d2d28bcbfea6591a282994062bd8d5eaa00636544292a0d268dca6e5da}) and compared it with the exact pinned mirror used for the model run. The corrected file contains 346,860 student--problem records, of which 283,105 contain a usable binary outcome and nonmissing skill identifier, from 4,163 learners. The mirror comprises 274,331 responses from 4,148 learners. All 4,148 mirror learner IDs occur in the corrected source; 15 additional learners occur only in the corrected source. Among the 4,148 shared learners, 3,331 have identical sequences of normalized (skill identifier, binary response) pairs when official records are ordered by numeric \texttt{order\_id}; 817 do not. The same 3,331 satisfy event-count and skill--response-multiset equality. Accordingly, \emph{exact whole-dataset parity is empirically disproved}, not merely unverified. The original mirror-derived predictive results remain correct for the pinned mirror, but cannot be represented as estimates from the corrected official 2009 release. This audit does not identify which upstream transformation rule caused the discrepancies. Although corrected-source learner and event keys can be checked directly, the mirror omits original student--problem event IDs, so a complete one-to-one event crosswalk and independently timestamped chronology are not recoverable. Numeric \texttt{order\_id} was used as a documented ordering convention, not as proof of real-world time order. The full aggregate audit, source hashes and code are archived in GitHub Actions run \href{https://github.com/devissaputra/knowledge_tracing_benchmark/actions/runs/39773920168}{39773920168}; individual student data are not redistributed.

\subsection{Officially linked 2015 independent-year source}
For the different-year replication we use the original ASSISTments 2015 Skill Builder CSV linked by the project's official dataset portal \citep{assistments2015}. The original file SHA-256 is \texttt{75e1f46131d5897a388c82411bc303}\texttt{2279899cb1f1e66a9043354d41234407a0}. It contains 708,631 records; 24,830 were excluded by the frozen missingness/validity filter, leaving 19,840 eligible learner IDs. The study constructs within-learner sequences in ascending numerical \texttt{log\_id} order, using \texttt{sequence\_id} as a 2015 skill-builder identifier; these identifiers are not the same construct as the 2009 mirror's \texttt{skill\_ids}. This different-year study requires new fitting on 2015 training learners and does not test a frozen 2009-trained model on 2015. The analysis source, hashes, sampling seeds, and both pilot and corrective-stage outcomes are archived in the repository's research branch. The 2015 result does not retroactively validate the 2009 mirror transformation.

\subsection{Partition unit, leakage checks, and model selection}
Learners are shuffled at seed 42 and divided into training (2,903), validation (622), and test (623) partitions. The runner first rejects duplicated \texttt{user\_id} sequence rows, then forbids learner overlap by partitioning whole serialized rows. It reports 194,951 training responses and 41,228 test responses. This check establishes no repeated \emph{learner identifier} across partitions, but does not verify the identity or chronological ordering of individual raw student--problem events. The validation set determines the BKT parameter choice and neural stopping epoch. Global and skill priors, PFA coefficients, and recurrent embeddings are learned from the training partition only. Every stateful model emits its prediction \emph{before} observing the target response.

\subsection{Model specifications}
\textbf{Global prior.} The binary response probability is the aggregate training correctness fraction. \textbf{Skill prior.} The response probability is the training correctness frequency for the queried skill, with a global fallback. \textbf{Fixed BKT.} Initial mastery, learning, guess, and slip probabilities are respectively 0.20, 0.15, 0.20, and 0.10. The prediction for a current response uses the previous posterior mastery state:
\begin{equation}
P(Y_t=1\mid H_{t-1})=m_{t-1}(1-s)+(1-m_{t-1})g.
\end{equation}
\textbf{Validation-tuned BKT.} The four parameters are selected from a fixed grid using validation log loss; the archived selection is initial mastery 0.40, learning 0.15, guess 0.30, and slip 0.20. \textbf{PFA.} A regularized logistic model uses a skill intercept and preceding learner--skill success/failure counts. \textbf{GRU.} A 64-dimensional recurrent hidden state processes previous response-conditioned skill tokens and is queried with the next skill embedding; seed-42 predictions for each learner's first response fall back to a training skill/global prior. Seeds 13, 42, and 73 were run for at most eight training epochs with validation-loss monitoring.

\textbf{Comparability caveat.} PFA counts only prior outcomes on the queried skill; the GRU can exploit the complete previous cross-skill sequence. The observed gap does not isolate the effect of nonlinearity or recurrence. The executed extension adds a PFA comparator with prior cross-skill success/failure counts, and longer GRU training. Because this comparator is still a compact summary rather than a full ordered-history model, the test does not isolate the effect of model architecture.

\subsection{Evaluation, uncertainty, and estimands}
ROC-AUC and average precision quantify discrimination and ranking. Brier score and log loss quantify probabilistic accuracy. Expected calibration error (ECE-10) compares confidence and observed response frequencies in ten probability bins, but is sample/bin-sensitive. A learner-first skill occurrence means that skill key has not appeared previously \emph{for that learner}; it is not necessarily unseen in training.

There are two distinct aggregation targets in the existing report. The primary table pools response-level losses, giving interaction-weighted estimates:
\begin{equation}
\widehat{B}_{\mathrm{event}}(f)=\frac{1}{N}\sum_{i=1}^{n}\sum_{t=1}^{T_i}(y_{it}-\hat p_{it}^{(f)})^2,\quad N=\sum_i T_i.
\end{equation}
The archived cluster bootstrap instead averages per-learner loss differences:
\begin{equation}
\widehat{\Delta}_{\mathrm{learner}}=\frac{1}{n}\sum_i\frac{1}{T_i}\sum_t\left[(y_{it}-\hat p_{it}^{(f)})^2-(y_{it}-\hat p_{it}^{(g)})^2\right].
\end{equation}
That is a defensible \emph{different} estimand. It must not be interpreted as an uncertainty interval for the event-pooled Brier difference. A paired learner-cluster bootstrap for the event-weighted difference resamples learners and recomputes the ratio of summed losses to summed response counts in each replicate. The executed supplement implements this correction, with 2,000 learner-resampling draws per comparison. Each draw recomputes the ratio of the sampled cluster loss sums to the sampled cluster response counts; the resulting percentile intervals are conditional on this one previously examined test partition.

\section{2009 Benchmark and Source Sensitivity}
\subsection{Overall held-out comparisons}
Table~\ref{tab:results} reports the results from a new full execution of the original study on the pinned real-data mirror. Reproduced original values agree with the archived record to within $4\times10^{-8}$ for ROC-AUC and much more closely for Brier score.
\begin{table}[ht]
\centering\small
\caption{Reproduced results on 41,228 responses from 623 learner-disjoint test learners (seed-42 original GRU). Lower Brier, log loss, and ECE are better.}
\label{tab:results}
\begin{tabular}{lrrrrr}\toprule
Model & ROC-AUC & AP & Brier & Log loss & ECE-10\\\midrule
Global prior & 0.5000 & 0.6747 & 0.2197 & 0.6313 & 0.0160\\
Skill prior & 0.6234 & 0.7669 & 0.2097 & 0.6096 & 0.0145\\
Fixed BKT & 0.6746 & 0.7806 & 0.2187 & 0.6360 & 0.1226\\
Validation-tuned BKT & 0.6722 & 0.7770 & 0.2024 & 0.5928 & 0.0279\\
PFA logistic & 0.6982 & 0.8074 & 0.1976 & 0.5840 & 0.0413\\
GRU (seed 42) & \textbf{0.7470} & \textbf{0.8439} & \textbf{0.1800} & \textbf{0.5382} & \textbf{0.0133}\\\bottomrule
\end{tabular}
\end{table}
The recorded GRU-minus-PFA Brier difference is approximately $-0.01765$ (an 8.9\% relative reduction with PFA as denominator), and ROC-AUC difference is $+0.04878$. These are \emph{descriptive differences} on one test partition. A separate \emph{exploratory} paired uncertainty calculation for the later strengthened comparison is reported below. In particular, the fixed-parameter BKT achieves higher AUC than the training-skill prior while having worse Brier score, illustrating why ranking and probability quality must be considered separately.

\begin{figure}[ht]\centering\includegraphics[width=.83\linewidth]{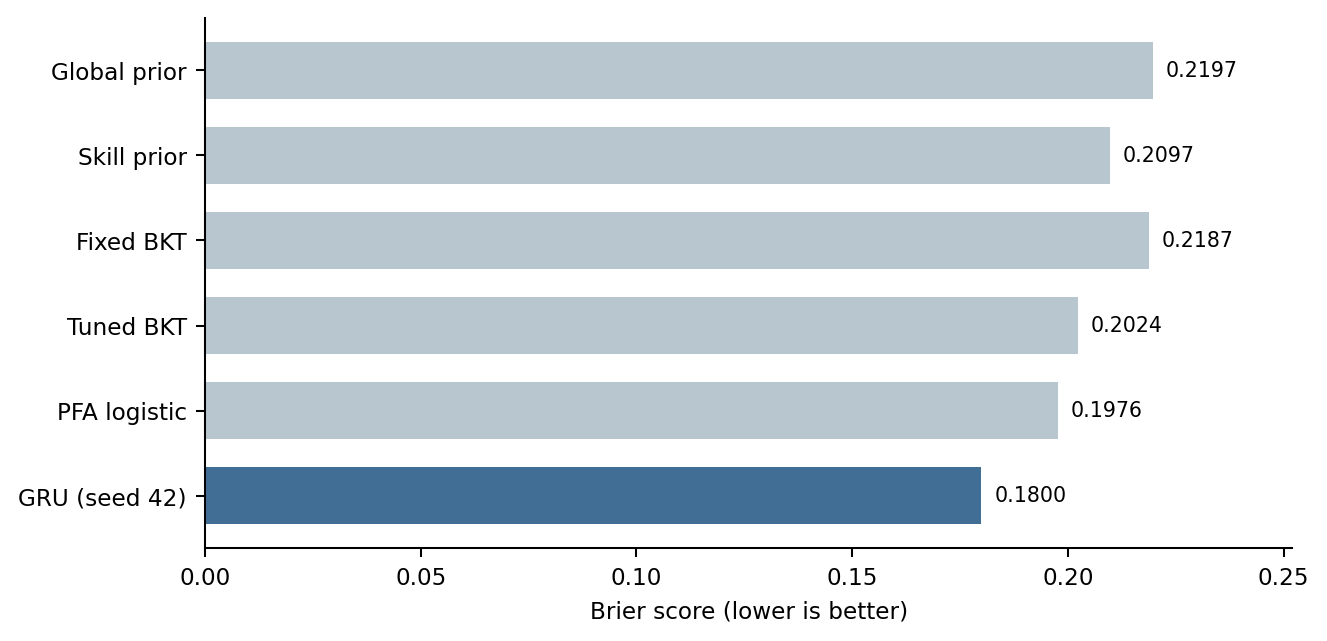}\caption{Reproduced response-pooled Brier scores for the original six benchmark models on the pinned real-data mirror.}\end{figure}

\subsection{First skill occurrence versus repeated occurrences}
\begin{table}[ht]\centering\small
\caption{Recorded PFA and seed-42 GRU performance by within-learner skill occurrence.}\begin{tabular}{lrrrrr}\toprule
Slice & N & PFA AUC & GRU AUC & PFA Brier & GRU Brier\\\midrule
First observed skill & 6,193 & 0.6499 & 0.7224 & 0.2244 & 0.2094\\
Repeated skill & 35,035 & 0.7068 & 0.7548 & 0.1929 & 0.1747\\\bottomrule
\end{tabular}\end{table}
The GRU has lower recorded Brier scores in both slices. However, this does not imply validated mastery estimation or a benefit for instructional decisions. Difficulty mix and number of available previous responses vary by slice.

\subsection{Random-seed variability and archived learner uncertainty}
Across the three prespecified GRU seeds the mean ROC-AUC is 0.74708 (sample SD 0.00083) and mean Brier score is 0.17975 (sample SD 0.00019). These quantify three optimization restarts on the same data split, not independent replications or evidence of robustness across populations.

The original run reproduces a \emph{mean per-learner} Brier difference between the seed-42 GRU and skill prior of $-0.02888$, with 95\% learner-block percentile interval $[-0.03439,-0.02365]$ for 623 learners. This interval must be linked to the learner-averaged quantity given above and cannot be relabelled as the pooled response-level comparison. The new cluster bootstrap for the interaction-pooled GRU--PFA comparison is presented separately below; its estimand must not be conflated with the original equal-learner-weighted interval.

\subsection{Independent execution reproducibility check}
The original end-to-end workflow was rerun in GitHub Actions (run \href{https://github.com/devissaputra/knowledge_tracing_benchmark/actions/runs/39655888620}{39655888620}) on 22 October 2026. The downloaded real-data file matched the archived SHA-256 \texttt{2053b7ea63f1d5ea953834bf82ed89543e1f11bae1e3e7535c2f9d7f49d4dda7}. Learner and response counts, learner partitions, original baseline Brier scores, and full-run AUCs matched archived results at reporting precision; the maximum original AUC drift was below $4\times10^{-8}$. Both code tests and the full data-validation job finished successfully. This is a computational reproduction of \emph{the same pinned third-party mirror}, not a parity audit against the official corrected source.

\subsection{Direct corrected-source parity test}
The corrected official release has 4,163 learners, including the 4,148 with sequences in the mirror. Normalizing the skill IDs and sorting the corrected file by numeric \texttt{order\_id} within each learner gives exact sequence agreement for 3,331 shared learner IDs; it fails for 817. Table~\ref{tab:parity} shows the necessary-condition comparison. Thus the frozen 2009 mirror results cannot be relabeled as corrected-release results. Ordered pair equality provides some external corroboration for unaffected learners, but it does not restore original problem-level identifiers to the mirror or prove that the \texttt{order\_id} convention captures the physical event chronology.
\begin{table}[ht]\centering\small
\caption{Official corrected 2009 skill-builder records compared with the pinned Atomi sequence mirror. Official labeled rows exclude records without a usable skill identifier.}
\label{tab:parity}
\begin{tabular}{lrr}\toprule
Measure & Corrected official & Pinned mirror\\\midrule
Unique learners & 4,163 & 4,148\\
Skill-labeled responses & 283,105 & 274,331\\
Shared learner IDs & \multicolumn{2}{c}{4,148}\\
Identical ordered skill--response sequences & \multicolumn{2}{c}{3,331 shared learners}\\
Different ordered sequences & \multicolumn{2}{c}{817 shared learners}\\\bottomrule
\end{tabular}
\end{table}

\subsection{Exploratory model and training-budget sensitivity}
To reduce the original information asymmetry, an enriched PFA baseline uses skill identity and both skill-specific and overall prior success/failure counts, with no access to the current response. The recurrent model still has richer ordered history and greater representational capacity; the comparison improves historical-information matching but does not equalize feature classes or prove an architectural mechanism. Extended GRU optimization used seeds 13, 42, and 73, at most 25 epochs, validation-loss early stopping and patience four. All three fits selected their best validation state before the cap. Table~\ref{tab:robust} summarizes the recorded out-of-sample scores \emph{on the same already-inspected test partition}.

\begin{table}[ht]\centering\small
\caption{Exploratory robustness results from real-data reruns. All rows use the same 623 test learners (41,228 responses); GRU settings are exploratory revisions after earlier test inspection.}
\label{tab:robust}
\begin{tabular}{lrrrr}\toprule
Model & ROC-AUC & Brier & Log loss & Best epoch\\\midrule
Standard PFA & 0.6982 & 0.1976 & 0.5840 & --\\
PFA + cross-skill counts & 0.7135 & 0.1929 & 0.5709 & --\\
GRU (seed 13, extended) & 0.7471 & 0.1793 & 0.5367 & 13\\
GRU (seed 42, extended) & 0.7491 & 0.1791 & 0.5357 & 13\\
GRU (seed 73, extended) & 0.7498 & 0.1789 & 0.5358 & 11\\\bottomrule
\end{tabular}
\end{table}

Relative to the stronger PFA, the extended seed-42 GRU has a response-pooled Brier improvement of $-0.013874$ (paired learner-cluster 95\% percentile interval $[-0.017231,-0.011049]$). Results for seeds 13 and 73 are $-0.013583$ [$-0.017004,-0.010911$] and $-0.014020$ [$-0.017418,-0.011123$], respectively. These bootstrap intervals (2,000 resamples of learners) describe conditional uncertainty \emph{on the fixed test partition after exploratory model changes}; they should not be interpreted as confirmatory evidence from an untouched test sample. The stronger PFA itself improves Brier score relative to the original PFA by $-0.004680$ [$-0.005900,-0.003537$], underscoring the effect of history-feature availability.

\begin{figure}[ht]\centering\includegraphics[width=.90\linewidth]{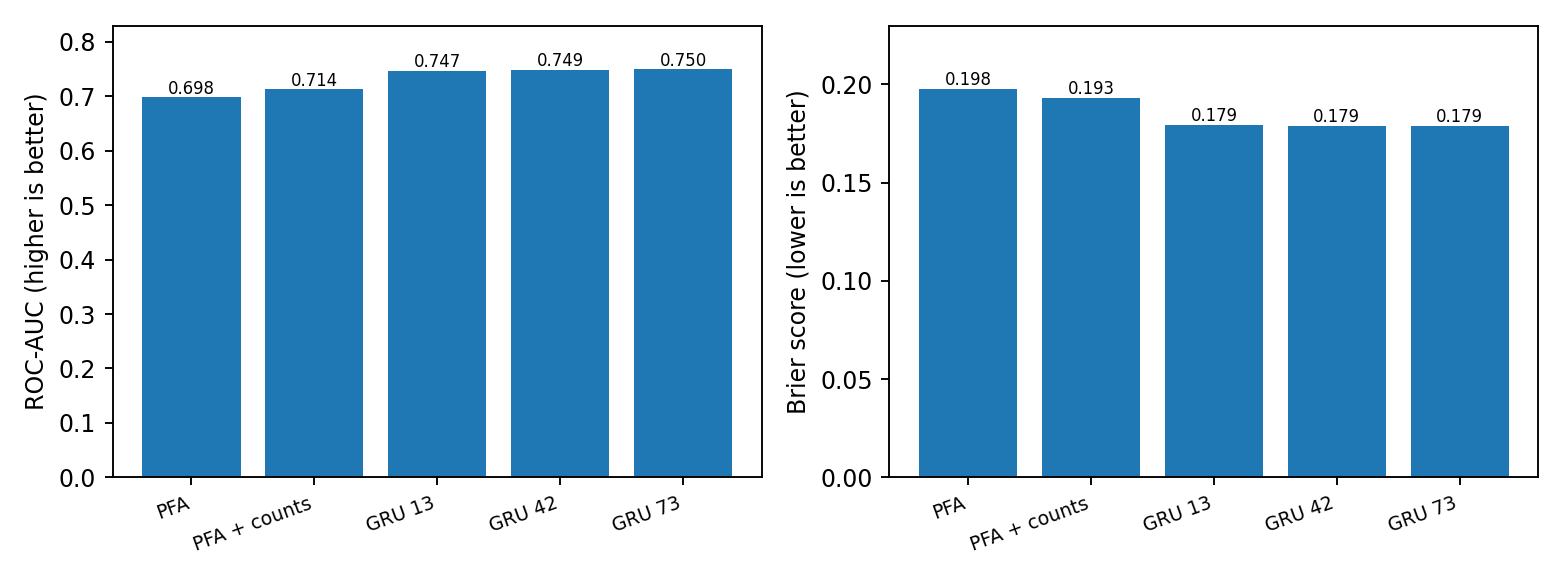}\caption{Exploratory robustness: real-data ROC-AUC and Brier score for PFA variants and three extended GRU seeds. This is a same-split sensitivity comparison, not an independent validation sample.}\end{figure}

\begin{figure}[ht]\centering\includegraphics[width=.83\linewidth]{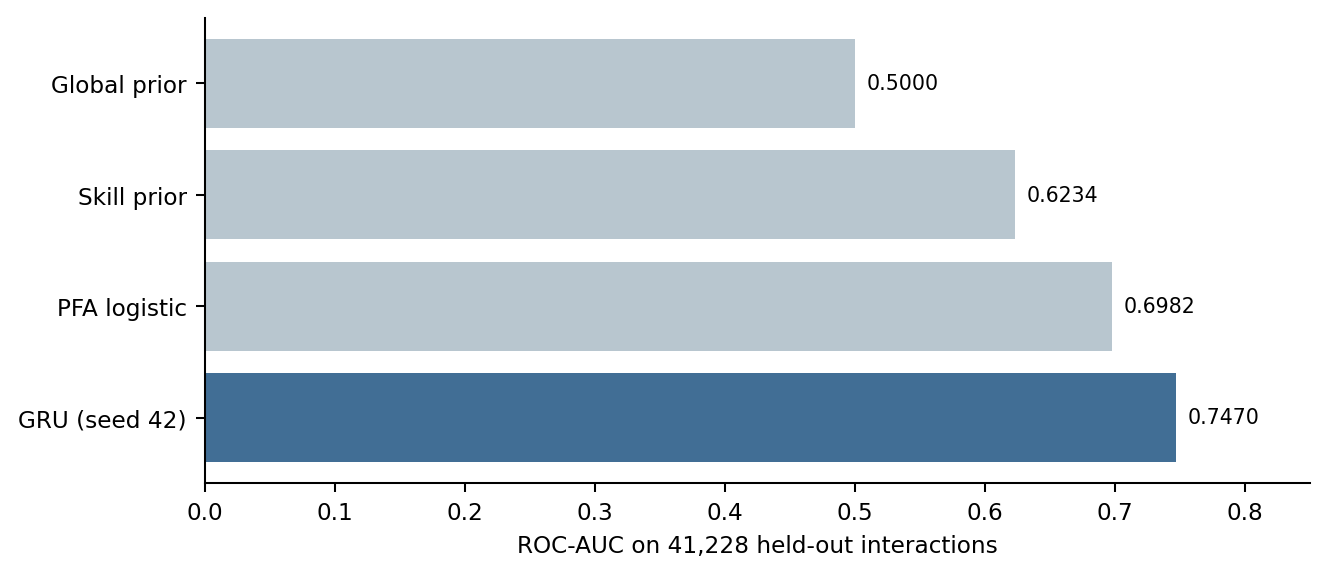}\caption{Reproduced test ROC-AUC across original models, shown descriptively for the fixed split.}\end{figure}

\subsection{Exploratory predictive sensitivity on corrected official 2009 data}
To determine how much the corrected source changes the predictive comparison, we ran a separate, prewritten \emph{exploratory} pipeline on the official corrected release rather than treating the mirror as equivalent. We excluded unlabeled skill records, normalized skill identifiers, ordered each learner's records by numeric \texttt{order\_id}, and used a whole-learner 70/15/15 split with seed 20261024. The enriched PFA retained its same-skill and across-skill history summaries, and the GRU retained prior ordered interactions, seed 42, 25-epoch maximum and validation-based stopping. Both models were refitted exclusively on corrected-source training learners. This is not a fully untouched confirmatory comparison, because the same 2009 academic setting and modeling direction were previously analyzed.

On 625 corrected-source test learners and 48,842 held-out interactions, the enriched PFA achieved ROC-AUC 0.6951, Brier 0.19675, and log loss 0.5809; the GRU achieved ROC-AUC 0.7324, Brier 0.18507, and log loss 0.5512. The paired interaction-weighted Brier difference (GRU minus PFA) was $-0.01168$ (95\% learner-cluster bootstrap percentile interval $[-0.01372,-0.00986]$, 2,000 resamples). The PFA optimizer converged in 930 iterations, and the GRU's best validation epoch was 13. These source-corrected figures are not directly numerically comparable with the 2009 mirror test split because its learner membership and preprocessing differ. The new analysis audits predictive sensitivity; it does not restore mirror/event parity or identify architectural causes. The aggregate code and outputs are archived in GitHub Actions run \href{https://github.com/devissaputra/knowledge_tracing_benchmark/actions/runs/39774414969}{39774414969}.

\section{Independent-Year ASSISTments 2015 Replication}
\subsection{Frozen sampling sequence and diagnostic correction}
The first 2015 analysis sampled 5,000 learners with seed 20261022, then assigned learners to 70\%/15\%/15\% train, validation, and test groups, comparing the enriched cross-skill PFA with the extended GRU. Although this initial pilot yielded a nominal GRU Brier advantage, the PFA \texttt{saga} optimizer reached its iteration limit and the declared convergence gate failed. The pilot is archived and \emph{not counted as clean confirmation}. Its test results must not be selectively omitted simply because they are provisional.

Before any second-stage test outcomes were inspected, a corrective protocol changed only the PFA numerical solver to \texttt{lbfgs} (\texttt{C=1}, maximum 2,000 iterations, tolerance $10^{-8}$) for the same regularized logistic model with skill intercepts, within-skill successes/failures, and pooled prior cross-skill successes/failures. It reconstructed and excluded every one of the 5,000 pilot learners, sampled 5,000 different learners (seed 20261023), and created learner-disjoint groups of 3,500 train, 750 validation, and 750 test learners. The GRU configuration remained fixed (seed 42; token embedding 32, skill query embedding 16, hidden size 64, Adam learning rate 0.002, up to 25 epochs, validation patience four). The GRU's best validation epoch was 7 of 11 executed. The enriched PFA optimizer converged in 484 iterations without convergence warnings. No observations from the first sample appear in the second sample.

\subsection{Untouched second-sample results and estimand}
The primary estimand is the interaction-pooled paired Brier loss difference, GRU minus enriched PFA, on the \emph{second} sample's final holdout. Two thousand learner-cluster bootstrap replicates (seed 20261023) resample whole learners and preserve the pooled-interaction weighting. A directional replication was counted only if the upper 95\% percentile interval endpoint was below zero and all provenance, disjointness, and convergence checks passed. This is a one-sided directional assessment using a displayed two-sided interval, not a preregistered general benchmark claim.

\begin{table}[ht]\centering\small
\caption{ASSISTments 2015 independent learner-cohort test (750 learners, 26,697 test responses). Higher AUC and lower Brier, log loss and ECE-10 are favorable. The second sample excludes all 5,000 learners from the first diagnostic pilot.}
\label{tab:cohort2015}
\begin{tabular}{lrrrr}\toprule
Model & ROC-AUC & Brier & Log loss & ECE-10\\\midrule
Enriched PFA (converged) & 0.68129 & 0.18695 & 0.55668 & 0.01694\\
Extended GRU (seed 42) & 0.69968 & 0.18227 & 0.54346 & 0.00820\\\bottomrule
\end{tabular}
\end{table}

The test correctness rate was 0.71697. The primary Brier difference was $-0.0046813$, with a 95\% learner-cluster percentile interval $[-0.0060730,-0.0033457]$. The prespecified \emph{second-stage} computational and directional gates passed, supported by GitHub Actions run \href{https://github.com/devissaputra/knowledge_tracing_benchmark/actions/runs/39754854426}{39754854426}. The first, nonconverged stage and its outcomes remain available in run \href{https://github.com/devissaputra/knowledge_tracing_benchmark/actions/runs/39754046492}{39754046492}. Because both 2015 experiments used the same source archive but \emph{nonoverlapping} learner samples, this is a fresh-sample, distinct-year predictive replication, not a demonstration of cross-institutional transport or strict information matching.

\section{2026 Matched-History-Budget Study on a New ASSISTments 2015 Cohort}
\subsection{Frozen stage-3 design and disjoint history budgets}
After completing the previous 2015 pilot and disjoint second-stage replication, we froze a separate methodological protocol \emph{before} evaluating a third learner sample (see \href{https://github.com/devissaputra/knowledge_tracing_benchmark/blob/research/kt-methodological-2026-history-budget/METHODOLOGICAL_2026_PROTOCOL.md}{frozen protocol}). The official 2015 source hash remained \texttt{75e1f46131d5897a388c82411bc3032279899cb1f1e66a9043354d41234407a0}. Using the original 19,840 eligible learner IDs and the published stage-1 and stage-2 random-seed sampling algorithms, we excluded \emph{all 10,000 learners} selected earlier. From the 9,840 remaining learners, we sampled a third 4,000 with seed 20261026, yielding 138,129 interactions. The learner-disjoint train/validation/test split contained 2,800/600/600 learners (the same seed).

Only responses after a learner's first recorded event are eligible targets, ensuring every evaluated response has at least one observed prior event. Consequently 18 of the 600 test learners contributed no history-eligible target, leaving \textbf{582 evaluable learners and 20,592 common next-response targets}. This distinction required correction of a downstream workflow assertion after the first computational run, with no changes to the model, split, data, features, training seeds or bootstrap. The initial failed release gate and corrected gate are preserved in the public workflow record.

For an event at position $t\geq1$, the permissible input comprises its current skill identifier and only the immediately preceding $\min(K,t)$ observed skill--correctness pairs, for a history budget $K\in\{1,20\}$. The current response is never an input. Count-based and neural models use the \emph{same observable information stream and identical response targets}. Training, validation and test learners remain disjoint; supervised fitting and standardization use training data only. Predictions on held-out learners are sequential online predictions, permitted to condition on \emph{earlier, already observed} responses within that learner. Such history access is not test-label leakage for the current event.

\subsection{Three fixed representations and descriptive estimands}
\textbf{Window-count PFA+:} Regularized logistic regression ($C=1$) with current-skill indicators and previous correct/incorrect counts for the queried skill and across all skills, computed strictly inside the last-$K$ window. \textbf{Order-aware logistic:} The identical count features plus the most recent prior correctness value, the latest result for the current skill (or absence flag) and lag to that skill within the window, all derivable from the permitted events. Both use LBFGS, standardized numeric features, 1,600-iteration ceiling and recorded convergence warnings. \textbf{Window-GRU:} Last-$K$ skill/outcome tokens enter a 48-unit hidden-state GRU with a separate 12-dimensional current-skill query embedding; token embedding 24, Adam learning rate 0.002, batch 256, up to 12 epochs, patience 3 on validation, seed 42. It sees no additional response history beyond the allowed K. Parameter counts and inductive biases remain unequal, so any gap cannot identify the isolated architecture effect.

Primary descriptive metrics are test Brier score, AUC, log loss and ECE-10. The principal model contrasts are paired \emph{interaction-weighted} GRU-minus-logistic Brier differences, with 2,000 learner-cluster percentile bootstrap resamples. These intervals describe sampling uncertainty conditional on one model fit and test split, without corrections for multiple models or budgets.

\subsection{Real-data results: order-aware logistic reverses the previous model ranking}
\begin{table}[t]
\centering\small
\caption{Same-target Stage-3 test performance at fixed prior-interaction budgets; $n=20{,}592$ responses from 582 evaluable learners at \emph{every} row. Lower Brier/log loss/ECE and higher AUC are better.}
\label{tab:budgetstudy}
\begin{tabular}{r l r r r r}
\toprule
$K$ & Model & Brier & AUC & Log loss & ECE-10\\
\midrule
1 & Window-count PFA+ & 0.17788 & 0.68444 & 0.53459 & 0.00932 \\
1 & Order-aware logistic & 0.17788 & 0.68444 & 0.53459 & 0.00932 \\
1 & Window-GRU & 0.18000 & 0.67091 & 0.54066 & 0.00578 \\
20 & Window-count PFA+ & 0.17667 & 0.69106 & 0.53159 & 0.00613 \\
20 & Order-aware logistic & 0.17389 & 0.70504 & 0.52409 & 0.00725 \\
20 & Window-GRU & 0.17514 & 0.69936 & 0.52715 & 0.01166 \\
\bottomrule
\end{tabular}
\end{table}
\begin{figure}[t]
\centering\includegraphics[width=.88\linewidth]{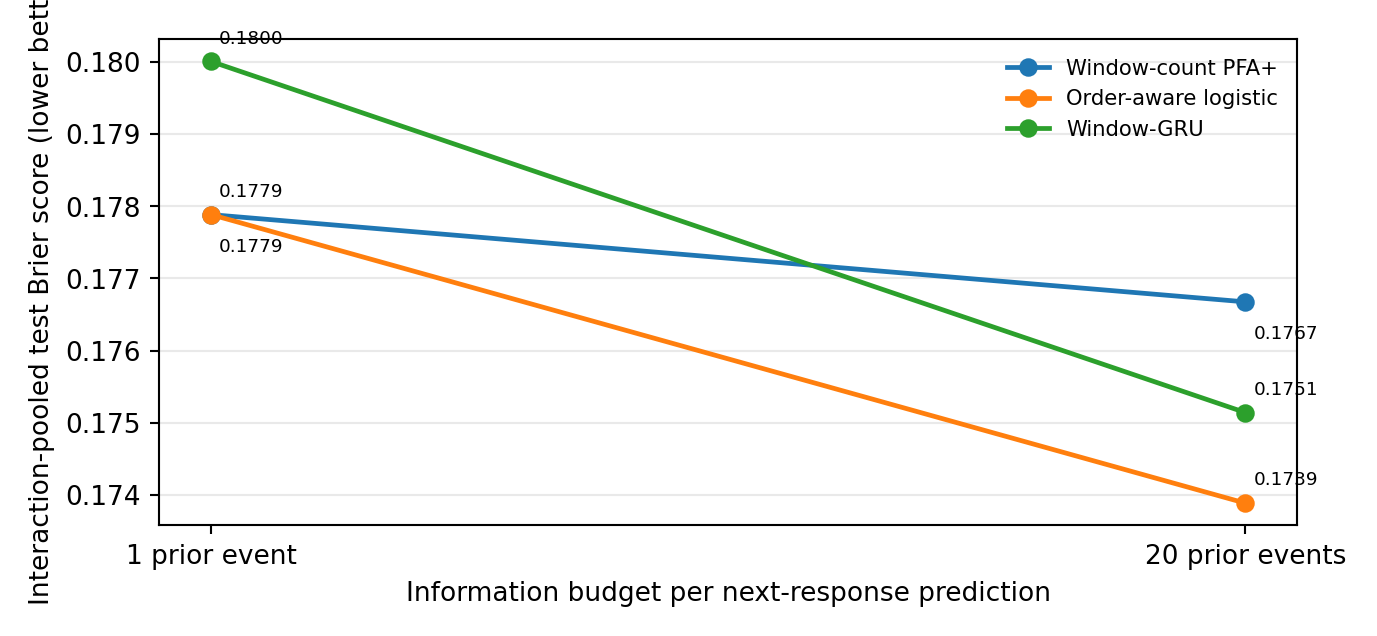}
\caption{Frozen Stage-3 budget sweep with shared train/validation/test learner partitions and exactly matching held-out prediction events. Error intervals for paired model differences are reported in text, not on the points.}
\label{fig:historybudget}
\end{figure}

The order-aware logistic model had the lowest recorded Brier score at both history budgets, although its $K=1$ advantage over the window-count model is immaterial at the reported precision. With only one previous response, count and order-aware models are numerically nearly identical, as expected: the single response already determines the recent-history summary. At 20 prior events, explicit order features improve the logistic Brier score from 0.17667 to 0.17389; the corresponding GRU is 0.17514. Moving from $K=1$ to $K=20$, Brier improves by \textbf{0.00121} for window-count, \textbf{0.00400} for order-aware logistic and \textbf{0.00486} for GRU. The original GRU advantage over standard PFA thus does not universally carry over when a transparent ordered-history comparator is given the same observable response window.

At $K=1$, GRU minus order-aware logistic had pooled Brier difference $+0.002121$ with a 95\% paired learner-cluster bootstrap interval $[+0.001198,+0.003013]$. Relative to window-count PFA+, the corresponding GRU difference was $+0.002121$, interval $[+0.001198,+0.003013]$.

At $K=20$, GRU minus order-aware logistic had pooled Brier difference $+0.001255$ with a 95\% paired learner-cluster bootstrap interval $[+0.000075,+0.002426]$. Relative to window-count PFA+, the corresponding GRU difference was $-0.001533$, interval $[-0.002761,-0.000279]$.

For both history budgets, the logistic solvers reported no convergence warnings. The GRU selected epoch 6 using validation loss. The preprocessing hash, nonoverlapping cohort, same-label target identity and numerical sanity gates passed. No family-wise significance claim or empirical generalization beyond the study's single held-out source is made. In particular, these confidence intervals are descriptive, not evidence that ordering \emph{causes} improved learning.

\section{Discussion and Threats to Validity}
\enlargethispage{1\baselineskip}
\textbf{Model ranking depends on the comparator.} On the third 2015 cohort, adding order-related features to logistic regression reduced Brier loss from 0.17667 to 0.17389 at $K=20$, below the GRU's 0.17514. At $K=1$, the two logistic specifications are nearly identical, and both score below the GRU. This does not show that logistic regression is intrinsically more accurate than recurrent models. Their feature engineering, parameter counts, optimization, and inductive biases differ. It does show that a GRU advantage against a coarse count baseline need not survive comparison with a stronger, transparent representation of the same history.

\textbf{Dataset identity matters.} The 2009 transformed mirror contains fewer labeled records than the corrected official source, and 817 shared learner sequences differ. Models trained and evaluated on the mirror therefore describe that specific transformation. The separate corrected-source results are not a paired estimate of the effect of preprocessing: their splits and target events differ. The mirror lacks original problem/event IDs needed for an exact row-level crosswalk.

\textbf{Inference remains conditional.} The 2009 exploratory revisions were made after inspecting the original holdout. In the first 2015 pilot, the PFA optimizer failed to converge; a revised solver was used on a disjoint second sample. The third protocol was fixed before its new learners were evaluated, but it was designed in light of the earlier experiments. Bootstrap intervals resample learners within the evaluated test split; they do not incorporate alternative training splits, tuning choices, multiple-testing adjustments, or shifts to other institutions.

The 2015 tests also use \texttt{sequence\_id} as the unit being predicted and order records by \texttt{log\_id}. Those coding and ordering decisions may not capture every aspect of the original learning activity. The matched-window experiment uses one split, one GRU seed, and two history lengths. Repeating it with additional seeds, modern attention-based baselines, and a genuinely different source would show whether the ranking is stable. Finally, accurate prediction of historical correctness does not establish that using any of these models improves instruction, fairness, or learning outcomes.

The GRU's additive linear head and the logistic model's explicit skill-specific history features differ, so the observed ranking cannot establish general architectural superiority. Learner separation and unique log identifiers cannot rule out repeated underlying question events. The 2015 sample covers selected high-volume skill builders, limiting generalizability. Cross-window improvements are descriptive, without separate uncertainty intervals.

\section{Summary of Findings}
The source audit rejects exact parity between the 2009 mirror and corrected release. The original six-model results reproduce on that mirror, but exploratory changes to model features or training cannot be treated as new confirmatory tests. The second 2015 sample reproduced the direction of the GRU--enriched-PFA Brier difference, conditional on the revised numerical solver and different model inputs. The third sample gives the clearest answer to the new question: with identical response targets and the same permitted event stream, an order-aware logistic model outperforms the tested GRU at both history lengths in Brier score. The contrast is small, especially at 20 events, and is specific to the implemented models and sample.

\section{Conclusion}
The transformed ASSISTments 2009 mirror is not equivalent to the corrected official source. The earlier GRU advantage on historical learner cohorts also depends on its comparison model: in a new 2015 evaluation with identical response targets, the order-aware logistic predictor achieved lower Brier loss than a compact GRU when both were given one or 20 preceding interactions. Allowing 20 previous responses lowered Brier loss for all three tested models relative to a one-response window, although ECE-10 did not improve for every model. The appropriate conclusion is that knowledge-tracing rankings are sensitive to source construction, history length, and representation. Stability across seeds, other datasets, and stronger neural baselines remains untested.

\section*{Data availability, computational evidence, and reproducibility boundary}
The official corrected 2009 and 2015 Skill Builder records can be obtained from the ASSISTments data portal \citep{assistmentsdata,assistments2015}. The publicly accessible \href{https://github.com/devissaputra/knowledge_tracing_benchmark}{analysis repository} contains the documented source checksums, preprocessing and modeling scripts, sampling protocols, machine-readable results, and software dependencies. Student-level records are not redistributed.

The recorded independent 2015 source-to-model reproduction, \href{https://github.com/devissaputra/knowledge_tracing_benchmark/actions/runs/40724144930}{Actions run 40724144930}, checked the official input SHA-256, rebuilt the nonoverlapping stage-3 cohort, refitted all six history-budget/model combinations, and recomputed paired learner-cluster intervals. The maximum absolute difference across the six Brier-score calculations was reported as $5.81\times 10^{-9}$. An independent \href{https://github.com/devissaputra/knowledge_tracing_benchmark/actions/runs/40724726591}{2009 source audit, run 40724726591}, recovered the 817 differing shared learner sequences. A separate \href{https://github.com/devissaputra/knowledge_tracing_benchmark/actions/runs/40731376499}{corrected-2009 model audit, run 40731376499}, reproduced the reported enriched-PFA/GRU contrasts within the recorded tolerance.

The original matched-history computation \href{https://github.com/devissaputra/knowledge_tracing_benchmark/actions/runs/40310733288}{(run 40310733288)} completed its numerical gates, but \emph{the workflow as a whole failed} because an automatic final Git push was rejected as non-fast-forward. It must not be cited as an entirely successful CI execution. After the workflow was repaired, \href{https://github.com/devissaputra/knowledge_tracing_benchmark/actions/runs/42347235331}{run 42347235331 on 7 November 2026} completed successfully, including calculation checks and artifact upload. These records establish reproducibility under the documented historical data and software conditions, not a causal architectural comparison, transportability, or improvement in teaching.

The present manuscript is supplied with its own semantically editable \LaTeX{} source and figures. Its editorial source edition is dated 7 November 2026 and should not be confused with older manuscript files retained in the GitHub repository.

\section*{Generative AI assistance and author accountability}
OpenAI's ChatGPT was used substantially in developing this work, including preparing and revising manuscript drafts, assisting with analysis code and technical documentation, formulating explanatory text about the methods and results, and checking \LaTeX{} presentation and source consistency. The reported numerical values were not accepted solely because a language model supplied them; the manuscript identifies the recorded analyses and independent computational checks supporting its principal results. Generative AI is not an author. Devis Wawan Saputra, the named human author, approved the scientific interpretations presented here and accepts responsibility for their accuracy, attribution, originality, disclosure, and any remaining limitations. Independent peer review and broad external replication have not been claimed.

\begingroup\small
\setlength{\itemsep}{0pt}

\endgroup
\end{document}